\documentclass{article}
\usepackage{ijcai21}

\usepackage{times}

\usepackage{soul}
\usepackage{url}
\usepackage[hidelinks]{hyperref}
\usepackage[utf8]{inputenc}
\usepackage[small]{caption}
\usepackage{graphicx}
\usepackage{amsmath}
\usepackage{booktabs}
\usepackage{amsmath, amssymb} 
\usepackage{cleveref} 
\usepackage{subcaption}  
\newcommand{\citet}[1]{\citeauthor{#1}~\shortcite{#1}}
\newcommand{\citep}{\cite}

\usepackage{bm} 
\newtheorem{thm}{Theorem}
\newtheorem{lem}{Lemma}

\title{Adversarial Training Makes Weight Loss Landscape\\ Sharper in Logistic Regression}

\author{
Masanori Yamada$^1$\footnote{Contact Author}\and
Sekitoshi Kanai$^2$\and
Tomoharu Iwata$^3$\and
Tomokatsu Takahashi$^1$\and
Yuki Yamanaka$^1$\and
Hiroshi Takahashi$^2$\And
Atsutoshi Kumagai$^{1,2}$\\
\affiliations
$^1$NTT Secure Platform Laboratories\\
$^2$NTT Software Innovation Center\\
$^3$NTT Communication Science Laboratories\\
\emails
\{masanori.yamada.cm, sekitoshi.kanai.fu, tomoharu.iwata.gy, tomokatsu.takahashi.wd, yuuki.yamanaka.kb, hiroshi.takahashi.bm, atsutoshi.kumagai.ht\}@hco.ntt.co.jp
}

\begin{document}
\maketitle
\begin{abstract}
Adversarial training is actively studied for learning robust models against adversarial examples. A recent study finds that adversarially trained models degenerate generalization performance on adversarial examples when their weight loss landscape, which is loss changes with respect to weights, is sharp. Unfortunately, it has been experimentally shown that adversarial training sharpens the weight loss landscape, but this phenomenon has not been theoretically clarified. Therefore, we theoretically analyze this phenomenon in this paper. As a first step, this paper proves that adversarial training with the $L_{\rm 2}$ norm constraints sharpens the weight loss landscape in the linear logistic regression model. Our analysis reveals that the sharpness of the weight loss landscape is caused by the noise aligned in the direction of increasing the loss, which is used in adversarial training. We theoretically and experimentally confirm that the weight loss landscape becomes sharper as the magnitude of the noise of adversarial training increases in the linear logistic regression model. Moreover, we experimentally confirm the same phenomena in ResNet18 with softmax as a more general case.
\end{abstract}

\section{Introduction}
Deep neural networks (DNNs) has been successfully used in wide applications such as image~\cite{he2016deep}, speech~\cite{wang2017residual}, and natural language processing~\cite{devlin2019bert}. Though DNNs have high generalization performance, they are vulnerable to adversarial examples, which are imperceptibly perturbed data to make DNNs misclassify data~\cite{szegedy2013intriguing}. For the real-world applications of deep learning, DNNs need to be made secure against this vulnerability.

Many methods to defend models against adversarial examples have been presented such as adversarial training~\cite{goodfellow2014explaining,kurakin2016adversarial,madry2017towards,zhang2019theoretically,Wang2020Improving}, adversarial detection~\cite{metzen2017detecting,huang2019model,NEURIPS2019_cbb6a3b8,liu2019detection},
defensive distillation~\cite{papernot2016distillation}, and random smoothing~\cite{cohen2019certified,pmlr-v119-yang20c,salman2019provably}.
Among them, adversarial training experimentally achieves high robustness~\cite{madry2017towards,zhang2019theoretically,Wang2020Improving,wu2020adversarial}.

Adversarial training trains models with adversarial examples. \citet{wu2020adversarial} experimentally shows that models obtain a high generalization performance in adversarial training when they have a flat weight loss landscape, which is the loss change with respect to the weight. However,~\citet{prabhu2019understanding} experimentally finds that adversarial training sharpens the weight loss landscape more than standard training, which is training without adversarial noise. Thus, the following question is important to answer: \textit{Does Adversarial Training Always Sharpen the Weight Loss Landscape?} If the answer is yes, adversarial training always has a larger generalization gap than standard training because the sharp weight loss landscape degrades the generalization performance.

In this paper, to answer this question theoretically, we focus on the logistic regression model with adversarial training as the first theoretical analysis step.  First, we use the definition of sharpness of the weight loss landscape that uses the eigenvalues of the Hessian matrix~\cite{keskar2016large}. Next, to simplify the analysis, we decompose the adversarial example into magnitude and direction. Finally, we show that the eigenvalues of the Hessian matrix in the weight loss landscape are proportional to the norm of the noise in the adversarial noise. As a results, we theoretically show that the weight loss landscape becomes sharper as the noise in the adversarial training becomes larger as shown in Theorem~\ref{thm:advsharpe}.
We experimentally confirmed Theorem~\ref{thm:advsharpe} on dataset MNIST2, which restricts MNIST to two classes. Moreover, we experimentally show that in multi-class classification with the nonlinear model (ResNet18) as a more general case, the weight loss landscape becomes sharper as the noise in the adversarial training becomes larger. Finally, to check whether the sharpness of the weight loss landscape is a problem specific to the adversarial example, we compare the weight loss landscapes of training on random noise with training on adversarially perturbed data in logistic regression. As a result, we confirmed theoretically and experimentally that the adversarial noise much sharpens the weight loss landscape more than the random noise. This is caused by noise always being added to adversarial examples in the direction that makes the loss worse. We conclude that the sharpness of the weight loss landscape needs to be reduced for reducing the generalization gap in adversarial training because the weight loss landscape becomes sharp in adversarial training, and the generalization gap becomes large.

Our contributions are as follows:
\begin{itemize}
  \item We show theoretically and experimentally that in logistic regression with the constrained $L_{\rm 2}$ norm, the weight loss landscape becomes sharper as the norm of the adversarial training noise increases.
  \item We show theoretically and experimentally that adversarial noise in the data space sharpens the weight loss landscape in logistic regression much more than random noise (random noise does not sharpen it extremely).
  \item We experimentally show that the larger norm of the noise of adversarial training makes the loss landscape sharper in the nonlinear model (ResNet18) with softmax. As a results, the generalization gap becomes larger as the norm of adversarial noise becomes large.
\end{itemize}


\section{Preliminary}
\subsection{Logistic Regression}
We consider a binary classification task with ${\bm x}\equiv(x_1,\dots,x_d) \in\mathbb{R}^{d}$ and $y\in\left\{-1,1\right\}$. A data point is represented as ${\bm x}^{n}$, where $n$ is the data index and its true label is $y^{n}$. A loss function of logistic regression is defined as
\begin{align}\label{eq:logisitc-regression-loss}
L\left(\bm{x},y,\bm{w}\right)&=\frac{1}{N}\sum_{n}\ell\left(\bm{x}^{n},y^{n},\bm{w}\right),\nonumber\\\ell\left(\bm{x}^{n},y^{n},\bm{w}\right)&\equiv\log\left(1+\exp\left(-y^{n}\bm{w}\cdot\bm{x}^{n}\right)\right),
\end{align}
where $N$ is the total number of data and ${\bm w}\equiv(w_1,\dots,w_d)\in\mathbb{R}^{d}$ is the training parameter of the model.

\subsection{Adversarial Example}
Adversarial example $\left({\bm x}'\right)^{n}={\bm x}^{n}+{\bm \eta}^{n}$ is defined as
\begin{align}
\bm{\eta}^{n}=\underset{\bm{\eta}^{n}\in\mathbb{B}_{{\rm p}}\left(\bm{x}^{n},\varepsilon\right)}{{\rm argmax}}\ell\left(\bm{x}^{n}+\bm{\eta}^{n},y^{n},\bm{w}\right),
\end{align}
where $\mathbb{B}_{\rm p}\left({\bm x}^{n},\varepsilon\right)$ is the region around ${\bm x}^{n}$ where the $L_{\rm{p}}$ norm is less than $\varepsilon$. Projected gradient decent (PGD)~\cite{madry2017towards}, which is a powerful adversarial attack, uses multi-step search as 
\begin{align}
\bm{x}^{n}\leftarrow\Pi_{\varepsilon}\left(\bm{x}^{n}+\lambda{\rm sign}\left(\frac{\partial}{\partial\bm{x}^{n}}\ell\left(\bm{x}^{n},y^{n},\bm{w}\right)\right)\right),
\end{align}
where $\lambda$ is step size and $\Pi_{\varepsilon}$ is projection to constrained space. Recently, since PGD can be prevented by gradient obfuscation, auto attack~\cite{croce2020reliable}, which tries various attacks, is used to more reliably evaluate robustness.
\subsection{Adversarial Training}
A promising way to defend the model against adversarial examples is adversarial training~\cite{goodfellow2014explaining,kurakin2016adversarial,madry2017towards}. Adversarial training learns the model with adversarial examples as 
\begin{align}
\min_{{\bm w}}\frac{1}{N}\sum_{n}\max_{{\bm \eta}^{n}\in \mathbb{B}_{\rm p}\left({\bm x}^{n},\varepsilon\right)}\ell\left(\bm{x}^{n}+\bm{\eta}^{n},y^{n},\bm{w}\right).
\end{align}
Since the adversarial training uses the information of the adversarial example, the adversarial training can improve the robustness of the model against the adversarial attack used for learning. In this paper, we describe models trained in standard training and adversarial training as clean and robust models, respectively. 

\subsection{Visualizing Loss Landscape}\label{sec:filter-normalization}
\citet{li2018visualizing} presented a filter normalization for visualizing the weight loss landscape.  The filter normalization works for not only standard training but also for adversarial training~\cite{wu2020adversarial}. The filter normalization with adversarial training visualizes the change of loss by adding noise to the weight as 
\begin{align}
g\left(\alpha\right)=\max_{{\bm \eta}^{n}\in \mathbb{B}_{\rm p}\left({\bm x}^{n},\varepsilon\right)}\frac{1}{N}\sum_{n}\ell\left({\bm x}^{n}+{\bm \eta}^{n},y^{n},{\bm w}+\alpha {\bm h}\right),
\end{align}
where $\alpha$ is the magnitude of noise and ${\bm h}\in\mathbb{R}^{d}$ is the direction of noise. The ${\bm h}$ is sampled from a Gaussian distribution and filter-wise normalized by 
\begin{align}
{\bm h}^{\left(l,m\right)}\leftarrow\frac{{\bm h}^{\left(l,m\right)}}{\left\Vert {\bm h}^{\left(l,m\right)}\right\Vert _{{\rm F}}}\left\Vert {\bm w}^{\left(l,m\right)}\right\Vert _{{\rm F}},
\end{align}
where ${\bm h}^{\left(l,m\right)}$ is the $m$-th filter at the $l$-th layer of ${\bm h}$ and $\left\Vert \cdot\right\Vert _{{\rm F}}$ is the Frobenius norm. We note that the Frobenius norm is equal to the $L_{\rm 2}$ norm in logistic regression because ${\bm w}$ can be regarded as $\bm{w}\in\mathbb{R}^{1\times d}$. This normalization is used to remove the \textit{scaling freedom} of weights for avoiding the fake loss landscape~\cite{li2018visualizing}. For instance, \citet{dinh2017sharp} exploit this scaling freedom of weights to build pairs of equivalent networks that have different apparent sharpnesses. The filter normalization absorbs the scaling freedom of the weights and enables the loss landscape to be visualized. The sharpness of the weight loss landscape strongly correlates with the generalization gap when this normalization is used~\cite{li2018visualizing}. Moreover, the sharpness of the weight loss landscape on adversarial training strongly correlates with a robust generalization gap when this normalization is used~\cite{wu2020adversarial}.

\section{Theoretical Analysis}
Since the robust model is trained by adding adversarial noise, the data loss landscape becomes flat. On the other hand, \citet{wu2020adversarial} experimentally confirmed that the weight loss landscape becomes sharp and the generalization gap becomes large in adversarial training. In this section, we theoretically show that the weight loss landscape becomes sharp in adversarial training.

\subsection{Definition of Weight Loss Landscape Sharpness}
Several weight loss landscape sharpness/flatness definitions have been presented in the study of generalization performance and loss landscape sharpness in deep learning~\cite{hochreiter1997flat,keskar2016large,chaudhari2017entropy}. In this paper, we use the simple definition, which defines flatness as the eigenvalues of the Hessian matrix. This definition is presented in~\cite{keskar2016large} and also used in~\cite{li2018visualizing}. For clarifying the relationship between the weight loss landscape and the generalization gap, we use filter normalization with weights. In other words, the definition of weight loss landscape sharpness is the eigenvalue of $\frac{\partial^{2}L}{\partial{w}_{i}\partial{w}_{j}}$ with the normalized weights. The larger the eigenvalue becomes, the sharper the weight loss landscape becomes.
\subsection{Main Results}
This section presents Theorem~\ref{thm:advsharpe}, which provides the relation between weight loss landscape and the norm of adversarial noise.

\begin{thm}\label{thm:advsharpe}
When the loss of the linear logistic regression model converges to the minimum for each data point ($\frac{\partial}{\partial w_{i}}l\left({\bm x}^{n}+{\bm \eta},y^{n},{\bm w}\right)=0$), the weight loss landscape become sharper for robust models trained with a large norm of adversarial noise that is the constrained in the $L_{\rm 2}$ norm.
\end{thm}

Let us prove Theorem~\ref{thm:advsharpe}. For the simple but special case of logistic regression, Eq.~(\ref{eq:logisitc-regression-loss}) can  be written with adversarial noise $\eta_{i}^{n}$ and angle $\theta^{n}$ as 

\scalebox{0.94}{\parbox{1.0\linewidth}{
\begin{align}
L_{\eta}=\!&\frac{1}{N}\!\sum_{n}\!\log\!\left(\!1\!+\!\exp\!\left(\!-y^{n}\!\bm{w}\!\cdot\!\left(\!\bm{x}^{n}\!+\!\bm{\eta}^{n}\!\right)\right)\right)\nonumber\\=\!&\frac{1}{N}\!\sum_{n}\!\log\!\left(\!1\!+\!\exp\!\left(\!-y^{n}\!\bm{w}\!\cdot\!\bm{x}^{n}\!+\!y^{n}\!\left\Vert\! \bm{w}\!\right\Vert \left\Vert\! \bm{\eta}^{n}\!\right\Vert \!\cos\!\theta^{n}\!\right)\!\right),\label{eq:adversarial-example}
\end{align}
}}
where $\left\Vert {\bm w}\right\Vert $ and $\left\Vert {\bm \eta}^{n}\right\Vert $ is the $L_{\rm 2}$ norm due to the inner product. Since an adversarial example must increase the loss, Lemma~\ref{lem:increasing-function} shows $y^{n}\cos\theta^{n}=1$.
\begin{lem}\label{lem:increasing-function}
$L_{\eta}$ is a monotonically increasing function of $\cos\theta^{n}$ when $y^{n}=1$, while $L_{\eta}$ is a monotonically decreasing function when $y^{n}=-1$.
\end{lem}
All the proofs of lemmas are provided in the supplementary material.

The gradient of loss with respect to weight $\frac{\partial L_{\eta}}{\partial\omega_{i}}$ is
\begin{equation}
\!\frac{1}{N}\!\sum_{n}\!\frac{\!\left(\!-y^{n}\!x_{i}^{n}\!+\!\frac{w_{i}}{\left\Vert \bm{w}\right\Vert }\!\left\Vert \!\bm{\eta}^{n}\!\right\Vert \!\right)\!\exp\!\left(\!-y^{n}\!\bm{w}\!\cdot\!\bm{x}^{n}\!+\left\Vert \bm{w}\right\Vert \left\Vert \bm{\eta}^{n}\!\right\Vert \!\right)}{\left(1+\exp\left(y^{n}\bm{w}\cdot\bm{x}^{n}-\left\Vert \bm{w}\right\Vert \left\Vert \bm{\eta}^{n}\right\Vert \right)\right)}.
\end{equation}
An optimal weight ${\bm w}^{\ast}$ of $L_{\eta}$ for each point satisfies $\frac{w_{i}^{\ast}}{\left\Vert {\bm w}^{\ast}\right\Vert }=\frac{y^{n}x_{i}^{n}}{\left\Vert {\bm \eta}^{n}\right\Vert }$. 
Next, we consider the Hessian matrix of loss on an optimal weight. The $(i,j)$-th element of the Hessian matrix is obtained as
\begin{align}\label{eq:hessian-eigenvalue}
\left.\frac{\partial^{2}L_{\eta}}{\partial\omega_{i}\partial\omega_{j}}\right|_{{\bm w}={\bm w}^{\ast}}=\frac{1}{2N}\sum_{n}\frac{\left\Vert {\bm \eta}^{n}\right\Vert }{\left\Vert {\bm w}^{\ast}\right\Vert }\left(\delta_{ij}-\frac{w_{i}^{\ast}w_{j}^{\ast}}{\left\Vert {\bm w}^{\ast}\right\Vert ^{2}}\right).
\end{align}
See the supplementary material for derivation. We consider the eigenvalues of the Hessian matrix. This matrix has a trivial eigenvector as
\begin{align}
\sum_{j}\left(\delta_{ij}-\frac{w_{i}^{\ast}w_{j}^{\ast}}{\left\Vert {\bm w}^{\ast}\right\Vert ^{2}}\right)v_{j}&=v_{i},\\\sum_{j}\left(\delta_{ij}-\frac{w_{i}^{\ast}w_{j}^{\ast}}{\left\Vert {\bm w}^{\ast}\right\Vert ^{2}}\right)w_{j}&=0,
\end{align}
where $v_j$ is the element of ${\bm v}$ which is an arbitrary vector orthogonal to ${\bm w}^{\ast}$. This matrix has eigenvalues $1$ and $0$. Since $\left(\delta_{ij}-\frac{w_{i}^{\ast}w_{j}^{\ast}}{\left\Vert {\bm w}^{\ast}\right\Vert ^{2}}\right) \in \mathbb{R}^{d \times d}$ is positive-semidefinite, $\frac{1}{2N}\sum_{n}\frac{\left\Vert {\bm \eta}^{n}\right\Vert }{\left\Vert {\bm w}^{\ast}\right\Vert }$ determines the sharpness of the weight loss landscape. Let $\left\Vert {\bm \eta}_{a}\right\Vert \geq\left\Vert {\bm \eta}_{b}\right\Vert $ be an adversarial example for perturbation strength $\varepsilon_{a}\geq\varepsilon_{b}$. The optimal weights in adversarial training with $\left\Vert {\bm \eta}_{a}\right\Vert $ and $\left\Vert {\bm \eta}_{b}\right\Vert $ are ${\bm w}_{a}^{\ast}$ and ${\bm w}_{b}^{\ast}$. The relation between the eigenvalues is as in 
\begin{align}\label{eq:norm-sharpness}
\frac{1}{2N}\sum_{n}\frac{\left\Vert {\bm \eta}_{a}\right\Vert }{\left\Vert {\bm w}_{a}^{\ast}\right\Vert }\geq\frac{1}{2N}\sum_{n}\frac{\left\Vert {\bm \eta}_{b}\right\Vert }{\left\Vert {\bm w}_{b}^{\ast}\right\Vert },
\end{align}
since the filter normalization makes the scale of the weights the same $\left\Vert {\bm w}_{a}^{\ast}\right\Vert \approx\left\Vert {\bm w}_{b}^{\ast}\right\Vert $ (In particularly in the case of logistic regression, this is natural because $\left\Vert {\bm w}\right\Vert _{{\rm F}}=\left\Vert {\bm w}\right\Vert$). Therefore, Theorem~\ref{thm:advsharpe} is proved from Eq.~(\ref{eq:norm-sharpness}). Moreover, since the eigenvalues are never negative, the results of adversarial training are always convex, which is a natural result.

\paragraph{Random Noise}
To clarify whether noise in data space sharpens the weight loss landscape is a phenomenon unique to adversarial training, we consider the weight loss landscape which trained on random noise. Let us assume a uniform distribution noise. We have the following theorem:

\begin{thm}\label{thm:random}
The logistic regression model trained with a uniform distribution noise, which is the constrained $L_{\infty}$ norm $\varepsilon$ as $\eta_{i}\sim U\left(-\varepsilon,\varepsilon\right)$. The eigenvalues of the Hessian matrix of this model converge to $0$ as the loss converges to minimum for each data point ($\frac{\partial}{\partial w_{i}}l\left({\bm x}^{n}+{\bm \eta},y^{n},{\bm w}\right)=0$). Furthermore, when the loss deviates slightly from the minimum, the Hessian matrix's eigenvalues become larger along with the norm of the random noise.
\end{thm}
This theorem shows that the weight loss landscape of the model trained with random noise is not as sharp as the weight loss landscape of the adversarial training model. 

Let us prove the Theorem~\ref{thm:random}. The derivative of the loss with arbitrary noise is
\begin{align}\label{eq:derivative-loss}
\frac{\partial L_{\eta}}{\partial w_{i}}&=-\frac{1}{N}\sum_{n}\frac{-y^{n}\left(g^{n}-1\right)x_{i}^{n}}{g^{n}},
\end{align}
where $g^{n}\equiv1+\exp\left(-y^{n}\bm{w}\cdot\left(\bm{x}^{n}+\bm{\eta}^{n}\right)\right)$. The $(i,j)$-th element of the Hessian matrix is obtained as
\begin{align}
\frac{\partial^{2}L_{\eta}}{\partial w_{i}\partial w_{j}}=\frac{1}{N}\sum_{n}\frac{\left(g^{n}-1\right)}{\left(g^{n}\right)^{2}}\left(x_{i}^{n}+\eta_{i}^{n}\right)\left(x_{j}^{n}+\eta_{j}^{n}\right).
\end{align}
We used $\left(y^{n}\right)^{2}=1$. Since the derivative of $\ell$ is zero on the optimal weight, the numerator of Eq.~(\ref{eq:derivative-loss}) becomes zero as $\kappa^{n}\equiv g^{n}-1\rightarrow0$. Thus, the eigenvalue of the Hessian matrix becomes $0$ on $g^{n}=1$. In other words, the weight loss landscape of the model trained with random noise converges closer to flat as the loss converges to the minimum. Let us consider the case where $\kappa^{n}$ is sufficiently small but takes a finite value independent of ${\bm \eta}$. Since we assume the uniform distribution noise, ${\bm x}$ and ${\bm \eta}$ uncorrelated and the mean of ${\bm \eta}$ is zero $\lim_{N\rightarrow\infty}\frac{1}{N}\sum_{n}\eta_{i}^{n}x_{j}^{n}=0$, and the variance of the noise is $\lim_{N\rightarrow\infty}\frac{1}{N}\sum_{n}\eta_{i}^{n}\eta_{j}^{n}=\frac{\varepsilon^{2}}{3}\delta_{ij}$. Thus, $\lim_{\kappa^{n}\rightarrow0}\frac{\partial^{2}L_{\eta}}{\partial w_{i}\partial w_{j}}$ in the large $N$  can be written as
\begin{align}
&\frac{1}{N}\sum_{n}\left(x_{i}^{n}x_{j}^{n}+\frac{\varepsilon^{2}}{3}\delta_{ij}\right)\lim_{\kappa^{n}\rightarrow0}\frac{\kappa^{n}}{\left(\kappa^{n}+1\right)^{2}}=0.
\end{align}
Thus, when $\kappa^{n}$ is sufficiently small but takes a finite value independent of ${\bm \eta}$, the weight loss landscape becomes sharp along with the norm of random noise. We note that since $\delta_{ij}$ is an element of the unit matrix, it increases the eigenvalue of the Hessian matrix. 
Considering Theorem~\ref{thm:advsharpe} and Theorem~\ref{thm:random}, the adversarial noise sharpens the weight loss landscape much more than the weight landscape on training with random noise. We have considered random noise constrained to the $L_{\infty}$ norm, but this statement holds if the mean of the random noise is zero and the variance of the random noise increases as the random noise norm increases. For example, considering Gaussian noise $\bm{\eta}\sim N\left(0,\sigma I\right)$, the same conclusion holds when the noise is $L_{{\rm 2}}$ norm constrained.



\section{Related Work}

\subsection{Weight Loss Landscape in Adversarial Training}
\citet{prabhu2019understanding} and~\citet{wu2020adversarial} experimentally studied the weight loss landscape in adversarial training. \citet{wu2020adversarial} demonstrated that the model with a flatter weight loss landscape has a smaller generalization gap and presented the Adversarial Weight Perturbation (AWP), which consequently achieves a more robust accuracy~\cite{wu2020adversarial}. \citet{prabhu2019understanding} reported that the robust model has a sharper loss landscape than the clean model. However, these studies did not theoretically analyze the weight loss landscape. Recently, \citet{liu2020loss} theoretically analyzed the loss landscapes in the robust model. The main topic of~\cite{liu2020loss} is a discussion of Lipschitzian smoothness in adversarial training, and their supplementary material contains a theoretical analysis of the weight loss landscape in logistic regression. The difference between the analysis of~\cite{liu2020loss} and our analysis is that~\citet{liu2020loss} compared the weight loss landscape of different $x$ positions in the single model ($L\left({\bm x},y,{\bm w}\right)$ and $L\left({\bm x}+{\bm \eta},y,{\bm w}\right)$), while we compare the weight loss landscape of the different models trained with different magnitudes of adversarial noise ($L\left({\bm x}+{\bm \eta}_{a},y,{\bm w}^{\ast}_{a}\right)$ and $L\left({\bm x}+{\bm \eta}_{b},y,{\bm w}^{\ast}_{b}\right)$, where ${\bm w}^{\ast}_{a}$ and ${\bm w}^{\ast}_{a}$ mean each optimal weights). Also, \citet{liu2020loss} used an approximation of the loss function as $L_{\eta}=\frac{1}{N}\sum_{n}y^{n}\log\left(1+\exp\left(-y^{n}{\bm w}\cdot{\bm x}^{n}+{\bm \eta}^{n}\right)\right)$ to simplify the problem. In contrast, we do not use an approximation of the loss function as Eq.~(\ref{eq:adversarial-example}). \citet{liu2020loss} derived that since the logistic loss is a monotonically decreasing convex function, adding noise in the direction of increasing the loss must increase the gradient, which leads to a sharp weight loss landscape.

\subsection{Weight Loss Landscape in Standard Training}
The relationship between weight loss landscape and generalization performance in deep learning has been theoretically and experimentally investigated~\cite{foret2020sharpness,Jiang*2020Fantastic,keskar2016large,dinh2017sharp}. In a large experiment evaluating 40 complexity measures by~\cite{Jiang*2020Fantastic}, measures based on the weight loss landscape had the highest correlation with the generalization error, and the generalization performance becomes better as the weight loss landscape becomes flatter. For improving the generalization performance by flattening the weight loss landscape, several methods have been presented, such as operating on diffused loss landscape~\cite{Mobahi2016}, local entropy regularization~\cite{chaudhari2017entropy}, and optimizer Sharpness-Aware Minimization (SAM)~\cite{foret2020sharpness}, which searches for a flat weight loss landscape. In particular, since a recently presented SAM~\cite{foret2020sharpness} is an improvement of the optimizer, it can be adapted to various methods and achieved a strong experimental result that updated the state-of-the-art for many datasets including CIFAR10, CIFAR100, and ImageNet. Since the weight loss landscape becomes sharper, and the generalization gap becomes larger in adversarial training than in standard training, we believe that finding a flat solution is more important in adversarial training than in standard training. 

\section{Experiments}
To verify the validity of Theorem~\ref{thm:advsharpe}, we visualize the sharpness of the weight loss landscape with various noise magnitudes in logistic regression in section~\ref{chap:logistic-exp}. Next, to investigate whether the sharpness of the weight loss landscape is a problem peculiar to adversarial training, we compare the training on random noise with the training on adversarial noise in section~\ref{chap:random-exp}. Finally, we visualize the weight loss landscape in a more general case (multi-class classification by softmax with deep learning) in section~\ref{chap:softmax-exp}. We also show the relationship between the weight loss landscape and the generalization gap. 

\paragraph{Experimental setup} We provide details of the experimental conditions in the supplementary material. We used three datasets: MNIST2, CIFAR10~\cite{krizhevsky2009learning}, and SVHN~\cite{netzer2011reading}. MNIST2 is explained the below subsection. We used PGD for adversarial training and robust generalization gap evaluation. For robustness evaluation, auto attack should be used. However, we did not use auto attack in our experiments since we are focusing on the generalization gap. For visualizing the weight loss landscape, we used filter normalization introduced in section~\ref{sec:filter-normalization}. The hyper-parameter settings for PGD were based on~\cite{madry2017towards} in MNIST2 and CIFAR10. Since there were no experiments using SVHN in~\cite{madry2017towards}, we used the hyper-parameter setting based on~\cite{wu2020adversarial} for SVHN. The $L_{2}$ norm of the perturbation $\varepsilon$ was set to $\varepsilon=\left\{0,0.2,0.4,0.6,0.8,1\right\} $ for MNIST2. The $L_{\infty}$ norm of the perturbation $\varepsilon$ was set to $\varepsilon=\left\{ \frac{0}{255},\frac{4}{255},\frac{8}{255},\frac{12}{255},\frac{16}{255},\frac{25.5}{255},\frac{51}{255},\frac{76.5}{255}\right\} $ for MNIST2 and $\varepsilon=\left\{ \frac{0}{255},\frac{4}{255},\frac{8}{255},\frac{12}{255},\frac{16}{255}\right\} $ for CIFAR10 and SVHN. For PGD, we updated the perturbation for $40$ iterations at training time and $100$ iterations at evaluation time on MNIST2. A step size of $L_{\rm 2}$PGD is $0.15\times\varepsilon$, while a step size of $L_{\infty}$PGD is $0.01$. For CIFAR10 and SVHN, we updated the perturbation for $7$ iterations at training time and $20$ iterations at evaluation time. A step size of $L_{\infty}$PGD is $\frac{2}{255}$ in CIFAR10, while a step size of $L_{\infty}$PGD is $\frac{1}{255}$ in SVHN. For random noise, we used uniformly distributed noise that is constrained in the $L_{\rm 2}$ or $L_{\infty}$ norm.

\begin{figure}[tb]
\centering
\includegraphics[width=0.81\linewidth]{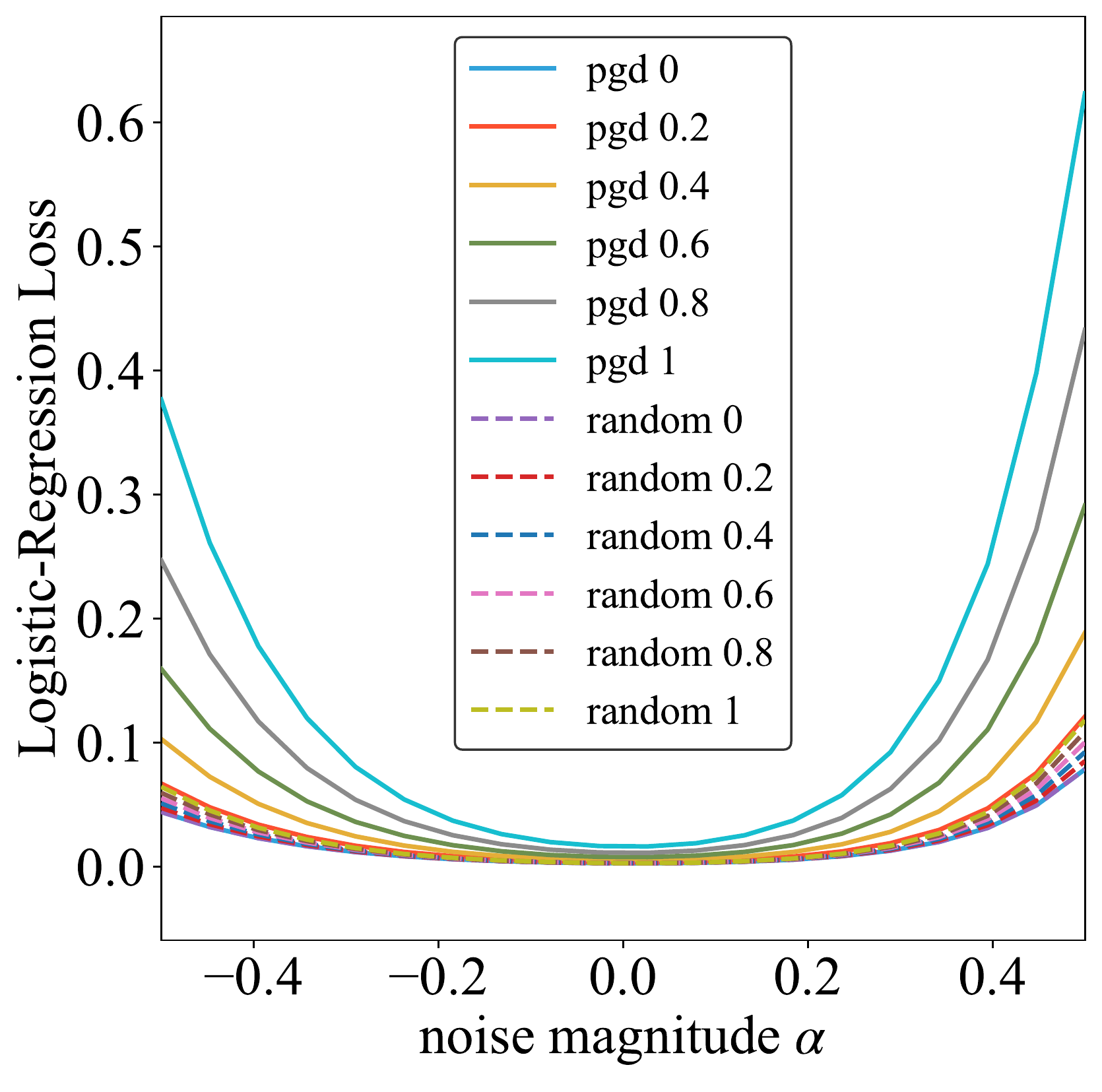}
\caption{Weight Loss Landscape against $L_{2}$PGD and $L_{2}$RANDOM in MNIST2}
\label{fig:landscape-mnist2-l2}
\centering
\includegraphics[width=0.81\linewidth]{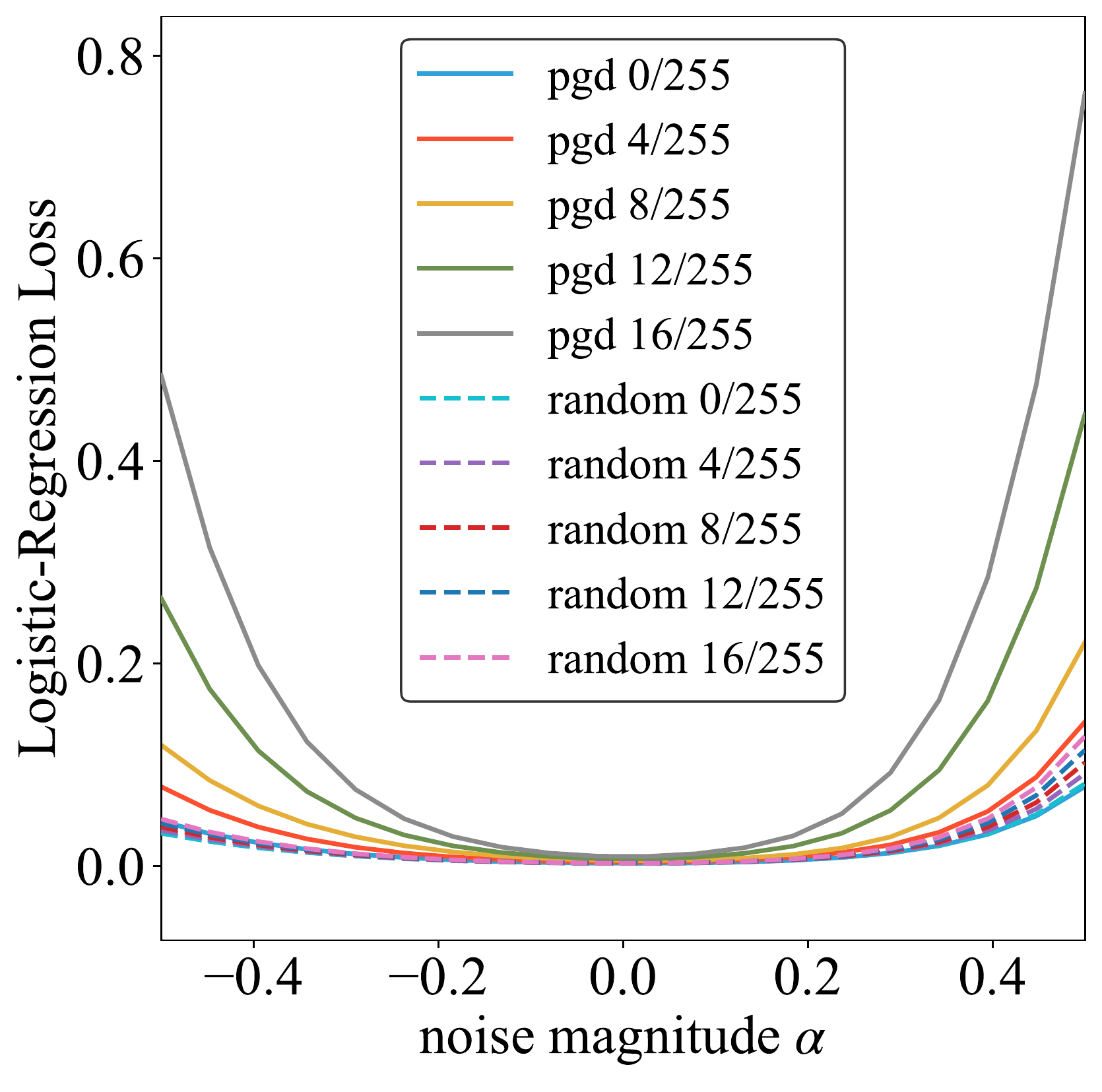}
\caption{Weight Loss Landscape against $L_{\infty}$PGD and $L_{\infty}$RANDOM in MNIST2}
\label{fig:landscape-mnist2}
\end{figure}

\begin{figure}[tb]
\begin{tabular}{cc}
\begin{minipage}{0.47\linewidth}
\includegraphics[width=1\linewidth]{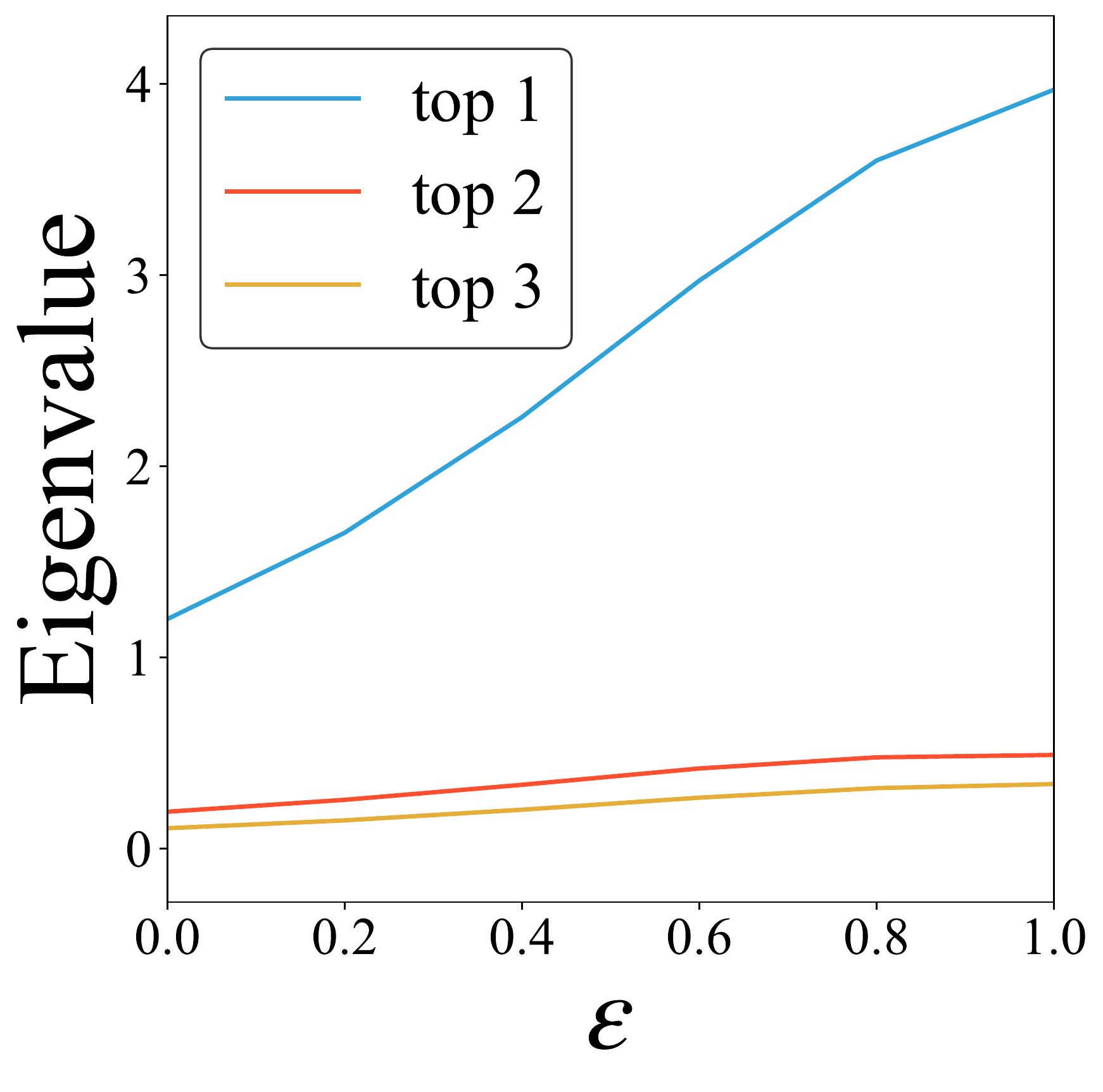}
\subcaption{$L_{\rm 2}$PGD}
\end{minipage}
\begin{minipage}{0.47\linewidth}
\includegraphics[width=1\linewidth]{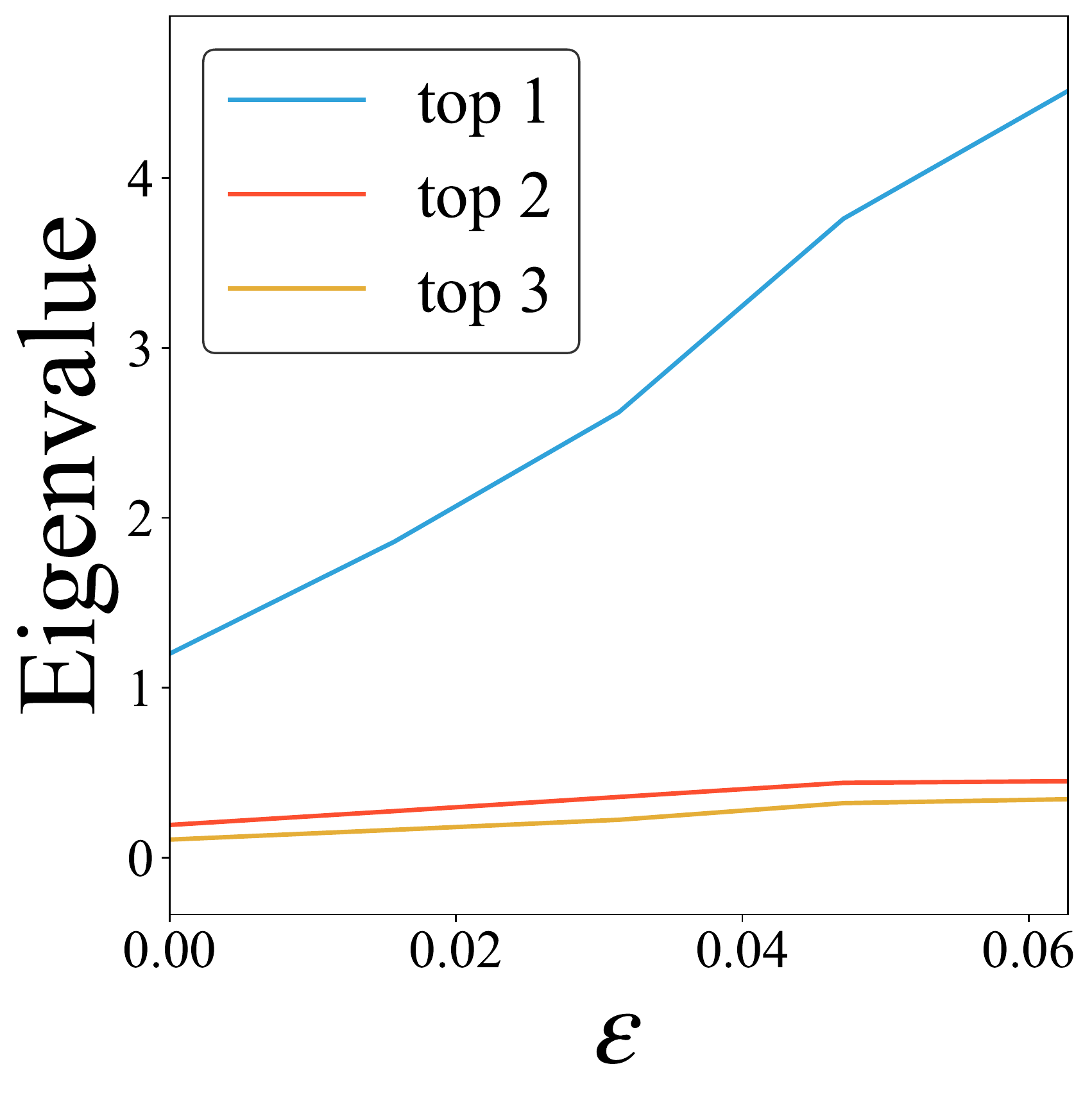}
\subcaption{$L_{\infty}$PGD}
\end{minipage}
\end{tabular}
\caption{Eigenvalue of Hessian matrix against PGD in MNIST2}
\label{fig:eigenvalue-mnist2-pgdl2-pgdlinf}
\end{figure}

\begin{table}[tb]
\centering
\caption{Robust accuracy against $L_{2}$PGD in MNIST2}
\begin{tabular}{crrr}
\toprule
\multicolumn{1}{c}{$\varepsilon$} & \multicolumn{1}{c}{train acc} & \multicolumn{1}{c}{test acc} & \multicolumn{1}{c}{gap} \\ \midrule
0.00 &      99.89 &     99.95 & -0.06 \\
0.20 &      99.79 &     99.95 & -0.16 \\
0.40 &      99.75 &     99.86 & -0.11 \\
0.60 &      99.76 &     99.76 & -0.00 \\
0.80 &      99.64 &     99.76 & -0.12 \\
1.00 &      99.49 &     99.62 & -0.13 \\
\bottomrule
\end{tabular}
\label{tab:mnist2-l2}
\end{table}

\begin{table}[tbh]
\centering
\caption{Robust accuracy against $L_{2}$RANDOM in MNIST2}
\begin{tabular}{crrr}
\toprule
\multicolumn{1}{c}{$\varepsilon$} & \multicolumn{1}{c}{train acc} & \multicolumn{1}{c}{test acc} & \multicolumn{1}{c}{gap} \\ \midrule
0.00 &      99.89 &     99.95 & -0.06 \\
0.20 &      99.89 &     99.95 & -0.06 \\
0.40 &      99.89 &     99.95 & -0.06 \\
0.60 &      99.89 &     99.95 & -0.06 \\
0.80 &      99.89 &     99.95 & -0.06 \\
1.00 &      99.88 &     99.95 & -0.07 \\
\bottomrule
\end{tabular}
\label{tab:mnist2-l2-random}
\end{table}

\begin{table}[tbh]
\centering
\caption{Robust accuracy against $L_{\infty}$PGD in MNIST2}
\begin{tabular}{rrrr}
\toprule
\multicolumn{1}{c}{$\varepsilon$} & \multicolumn{1}{c}{train acc} & \multicolumn{1}{c}{test acc} & \multicolumn{1}{c}{gap} \\ \midrule
    0/255 & 99.89 & 99.91 & -0.02 \\
  4/255 &      99.79 &     99.95 & -0.16 \\
  8/255 &      99.74 &     99.86 & -0.12 \\
 12/255 &      99.79 &     99.76 &  0.02 \\
 16/255 &      99.72 &     99.76 & -0.05 \\
25.5/255 & 99.46 & 99.67 & -0.21 \\
51/255 & 97.84 & 98.53 & -0.70 \\
76.5/255 & 95.40 & 96.64 & -1.24 \\
\bottomrule
\end{tabular}
\label{tab:mnist2}
\end{table}

\begin{table}[tbh]
\centering
\caption{Robust accuracy against $L_{\infty}$RANDOM in MNIST2}
\begin{tabular}{rrrr}
\toprule
\multicolumn{1}{c}{$\varepsilon$} & \multicolumn{1}{c}{train acc} & \multicolumn{1}{c}{test acc} & \multicolumn{1}{c}{gap} \\ \midrule
  0/255 &      99.89 &     99.91 & -0.02 \\
  4/255 &      99.89 &     99.91 & -0.02 \\
  8/255 &      99.89 &     99.91 & -0.02 \\
 12/255 &      99.89 &     99.91 & -0.02 \\
 16/255 &      99.89 &     99.91 & -0.02 \\
\bottomrule
\end{tabular}
\label{tab:mnist2-random}
\end{table}

\subsection{Binary Logistic Regression}\label{chap:logistic-exp}
We conducted experiments on image dataset MNIST~\cite{lecun1998gradient}, which is well known for its adversarial example settings. Since the linear logistic regression model did not perform well in classifying the two-class CIFAR10, we evaluated on MNIST. To experiment with logistic regression for binary classification, we created a two-class dataset MNIST2 from MNIST. We made MNIST2 using only MNIST class 0 and class 1. \Cref{fig:landscape-mnist2-l2,fig:landscape-mnist2} show the weight loss landscape with various noise magnitudes of the $L_{2}$ and $L_{\infty}$ norm in logistic regression. We can confirm that the weight loss landscape becomes sharp as the noise magnitude increases. For the sake of clarity, we have excluded the ranges with large $\varepsilon=\left\{\frac{25.5}{255},\frac{51}{255},\frac{76.5}{255}\right\}$ in $L_{\infty}$ from Fig.~\ref{fig:landscape-mnist2}. The results for the large $\varepsilon=\left\{ \frac{0}{255},\frac{25.5}{255},\frac{51}{255},\frac{76.5}{255}\right\}$ in $L_{\infty}$ norm, which is often used in the 10-class classification of MNIST, are included in the supplementary material. The results are similar: the larger the noise magnitude, the sharper the weight loss landscape. Table~\ref{tab:mnist2-l2} also shows that the absolute value of the generalization gap is larger in adversarial training than in standard training ($\varepsilon=0$). The test robust accuracy is larger than the training accuracy because of early stopping in the test robust accuracy~\cite{rice2020overfitting}, but we emphasize that the training accuracy and test accuracy diverge. 
In Table.~\ref{tab:mnist2}, the relationship between the generalization gap and $\varepsilon$ is difficult to understand because the experiment was designed with a small noise to achieve training loss becomes zero. However, where $\varepsilon$ is a large range in Tab.~\ref{tab:mnist2}, the absolute value of the generalization gap increases as $\varepsilon$ increases in the experiment.  

\begin{figure}[tb]
\begin{tabular}{cc}
\begin{minipage}{0.47\linewidth}
\includegraphics[width=1\linewidth]{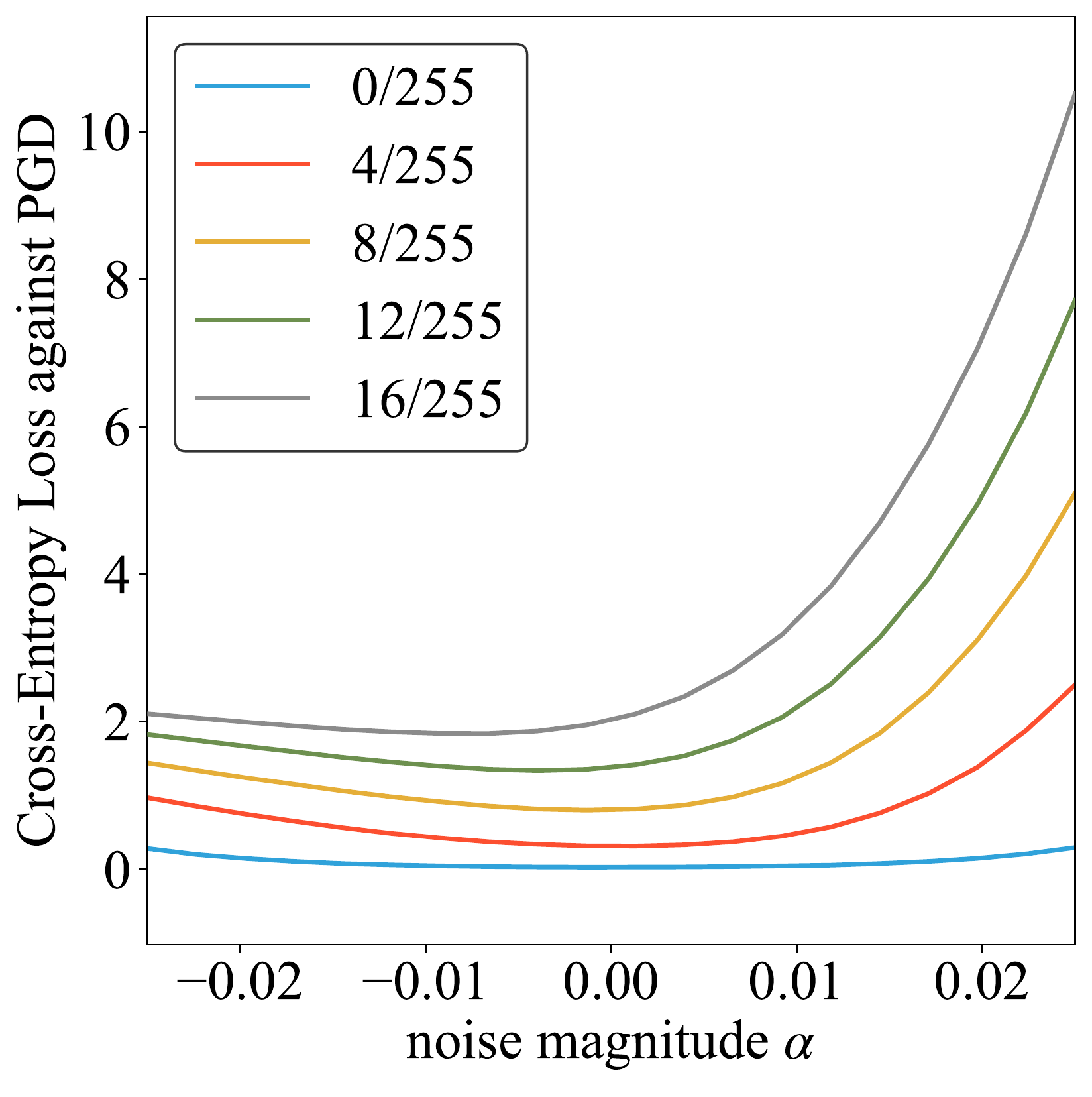}
\subcaption{CIFAR10}
\end{minipage}

\begin{minipage}{0.47\linewidth}
\includegraphics[width=1\linewidth]{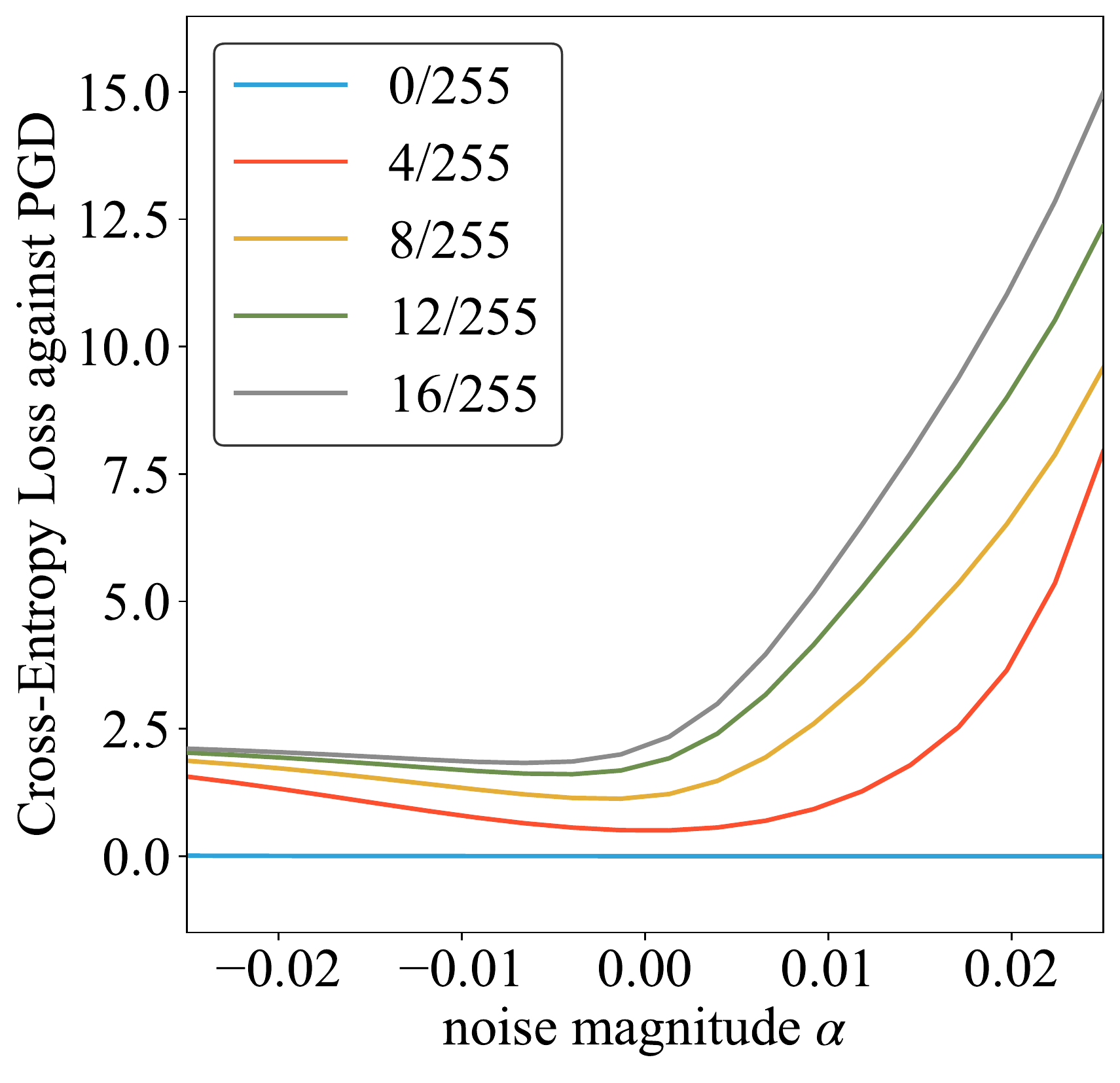}
\subcaption{SVHN}
\end{minipage}
\end{tabular}
\caption{Weight Loss Landscape against $L_{\infty}$PGD in ResNet18}
\label{fig:landscape-CIFAR10-SVHN}
\end{figure}

\begin{table}[tbh]
\centering
\caption{Robust accuracy against $L_{\infty}$PGD in CIFAR10}
\begin{tabular}{@{}rrrr@{}}
\toprule
\multicolumn{1}{c}{$\varepsilon$} & \multicolumn{1}{c}{train acc} & \multicolumn{1}{c}{test acc} & \multicolumn{1}{c}{gap} \\ \midrule
0/255 & 99.00 & 93.89 & 5.11 \\
4/255 & 81.90 & 69.73 & 12.17 \\
8/255 & 63.81 & 50.98 & 12.83 \\
12/255 & 50.83 & 36.69 & 14.14 \\
16/255 & 47.56 & 25.61 & 21.95 \\ \bottomrule
\end{tabular}
\label{tab:cifar10}
\end{table}

\begin{table}[tbh]
\centering
\caption{Robust accuracy against $L_{\infty}$PGD in SVHN}
\begin{tabular}{@{}rrrr@{}}
\toprule
\multicolumn{1}{c}{$\varepsilon$} & \multicolumn{1}{c}{train acc} & \multicolumn{1}{c}{test acc} & \multicolumn{1}{c}{gap} \\ \midrule
   0/255 &     100.00 &     96.13 &   3.87 \\
   4/255 &      83.53 &     73.51 &  10.01 \\
   8/255 &      70.96 &     51.82 &  19.14 \\
  12/255 &      64.58 &     36.93 &  27.65 \\
  16/255 &      60.93 &     29.59 &  31.34 \\
\bottomrule
\end{tabular}
\label{tab:svhn}
\end{table}

\paragraph{Eigenvalue of Hessian matrix}
We analyze the eigenvalue of the Hessian matrix for confirming Eq.~(\ref{eq:hessian-eigenvalue}). The results of the linear logistic regression model in MNIST2 are shown. Figure~\ref{fig:eigenvalue-mnist2-pgdl2-pgdlinf} shows the top three eigenvalues of the model trained with $L_{\rm 2}$ PGD and $L_{\infty}$ PGD with different epsilons. These figures show that the eigenvalues are nearly linear with respect to epsilon, as in our theoretical analysis. We estimated the eigenvalues without filter normalization, but we checked that the norm of weight does not change significantly with each epsilon.

\paragraph{Comparison with Random Noise}\label{chap:random-exp}
To clarify whether noise in data space sharpens the weight loss landscape is a phenomenon unique to adversarial training, we compare the weight loss landscape of learning with random noise and learning with adversarial noise (adversarial training with PGD). We used a logistic regression model in MNIST2, and random noise was generated with a uniform distribution constrained to the $L_{{\rm 2}}=\left\{ 0,0.2,0.4,0.8,1\right\} $ and $L_{\infty}=\left\{ 0,\frac{4}{255},\frac{8}{255},\frac{12}{255},\frac{16}{255}\right\} $, and then constrained to fit in the range $\left[0,1\right]$ of the normalized image along with the image. \Cref{fig:landscape-mnist2-l2,fig:landscape-mnist2} compare the weight loss landscape in adversarial noise with random noise in the $L_{{\rm 2}}$ and $L_{\infty}$ norm, respectively. As seen in our theoretical analysis, these figures show that the adversarial noise sharpens the weight loss landscape much more than the weight landscape on training with random noise for both the $L_{{\rm 2}}$ and $L_{\infty}$ norms. As a result, the generalization gap in random noise is smaller than that in the adversarial noise as shown in \Cref{tab:mnist2-l2-random,tab:mnist2-random}.

\subsection{Multi-class Classification}\label{chap:softmax-exp}
In the more general case as multi-class classification using softmax with residual network~\cite{he2016deep}, we confirm that the weight loss landscape becomes sharper in adversarial training as well as logistic regression. We use ResNet18~\cite{he2016deep} with softmax in CIFAR10 and SVHN. Figure~\ref{fig:landscape-CIFAR10-SVHN} shows the weight loss landscape in $\varepsilon=\left\{ \frac{0}{255},\frac{4}{255},\frac{8}{255},\frac{12}{255},\frac{16}{255}\right\} $. We confirmed that the weight loss landscape becomes sharper as the noise magnitude of adversarial training becomes larger. \Cref{tab:cifar10,tab:svhn} also shows that the generalization gap becomes larger in most cases as the magnitude of noise becomes larger.

\section{Conclusion and Future work}
In this paper, we show theoretically and experimentally that the weight loss landscape becomes sharper when the noise in adversarial training is strong with the linear logistic regression model. In linear logistic regression, we also showed that not all data space noises make the landscape extremely sharp, but adversarial examples make the weight loss landscape extremely sharp, both theoretically and experimentally. The theoretical analysis in a more general nonlinear model (such as a residual network) with softmax is future work. To motivate future work, we experimentally showed that the weight loss landscape becomes sharper when the noise of adversarial training becomes larger. We conclude that the sharpness of the weight loss landscape needs to be reduced for reducing the generalization gap in adversarial training
because the weight loss landscape becomes sharp in adversarial training, and the generalization gap becomes large.

\appendix

\section{Proof of Lemma~1}
We show that loss is monotonically increasing for $\cos\theta^{n}$ when $y^{n}=1$ and monotonically decreasing when $y^{n}=-1$.
\scalebox{0.96}{\parbox{1.0\linewidth}{
\begin{align}
&\frac{\partial}{\partial\left(\cos\theta^{n}\right)}\log\left(1+\exp\left(-y^{n}\bm{w}\cdot\bm{x}^{n}+y^{n}\left\Vert \bm{w}\right\Vert \left\Vert \bm{\eta}^{n}\right\Vert \cos\theta^{n}\right)\right)\nonumber\\&=\frac{\frac{\partial}{\partial\left(\cos\theta^{n}\right)}\left(\exp\left(-y^{n}\bm{w}\cdot\bm{x}^{n}+y^{n}\left\Vert \bm{w}\right\Vert \left\Vert \bm{\eta}^{n}\right\Vert \cos\theta^{n}\right)\right)}{1+\exp\left(-y^{n}\bm{w}\cdot\bm{x}^{n}+y^{n}\left\Vert \bm{w}\right\Vert \left\Vert \bm{\eta}^{n}\right\Vert \cos\theta^{n}\right)}\nonumber\\&=\frac{y^{n}\left\Vert \bm{w}\right\Vert \left\Vert \bm{\eta}^{n}\right\Vert \left(\exp\left(-y^{n}\bm{w}\cdot\bm{x}^{n}+y^{n}\left\Vert \bm{w}\right\Vert \left\Vert \bm{\eta}^{n}\right\Vert \cos\theta^{n}\right)\right)}{1+\exp\left(-y^{n}\bm{w}\cdot\bm{x}^{n}+y^{n}\left\Vert \bm{w}\right\Vert \left\Vert \bm{\eta}^{n}\right\Vert \cos\theta^{n}\right)}\nonumber
\end{align}
}}
Since the exponential function always has a positive value, Lemma~1 is proved.

\section{Hessian matrix on the optimal weight}
Compute the Hessian matrix of the adversarial training model of logistic regression. The loss function is 
\begin{align}
L\left(\bm{x},y,\bm{w}\right)&=\frac{1}{N}\sum_{n}\ell\left(\bm{x}^{n},y^{n},\bm{w}\right),\nonumber\\\ell\left(\bm{x}^{n},y^{n},\bm{w}\right)&\equiv\log\left(1+\exp\left(-y^{n}\bm{w}\cdot\bm{x}^{n}\right)\right)\nonumber.
\end{align}
The Hessian matrix of loss along weight $\frac{\partial^{2}L_{\eta}}{\partial\omega_{i}\partial\omega_{j}}$ is
\begin{align}
&\frac{1}{N}\sum_{n}\left(\frac{\frac{\partial}{\partial\omega_{j}}\left(-\text{\ensuremath{y^{n}}}x_{i}^{n}+\frac{w_{i}}{\left\Vert {\bm w}\right\Vert }\left\Vert {\bm \eta}^{n}\right\Vert \right)}{\left(1+\exp\left(\sum_{i}\text{\ensuremath{y^{n}}}w_{i}x_{i}^{n}-\left\Vert {\bm w}\right\Vert \left\Vert {\bm \eta}^{n}\right\Vert \right)\right)}\right.\nonumber\\&+\left(-\text{\ensuremath{y^{n}}}x_{i}^{n}+\frac{w_{i}}{\left\Vert {\bm w}\right\Vert }\left\Vert {\bm \eta}^{n}\right\Vert \right)\nonumber\\&\left.\frac{\partial}{\partial\omega_{j}}\frac{1}{\left(1+\exp\left(\sum_{i}\text{\ensuremath{y^{n}}}w_{i}x_{i}^{n}-\left\Vert {\bm w}\right\Vert \left\Vert {\bm \eta}^{n}\right\Vert \right)\right)}\right)\nonumber\\&=\frac{1}{N}\sum_{n}\left(\frac{\frac{\left\Vert {\bm \eta}^{n}\right\Vert }{\left\Vert {\bm w}\right\Vert }\left(\delta_{ij}-\frac{w_{i}w_{j}}{\left\Vert {\bm w}\right\Vert ^{2}}\right)}{\left(1+\exp\left(\sum_{i}\text{\ensuremath{y^{n}}}w_{i}x_{i}^{n}-\left\Vert {\bm w}\right\Vert \left\Vert {\bm \eta}^{n}\right\Vert \right)\right)}\right.\nonumber\\
&+\left(-\text{\ensuremath{y^{n}}}x_{i}^{n}+\frac{w_{i}}{\left\Vert {\bm w}\right\Vert }\left\Vert {\bm \eta}^{n}\right\Vert \right)\nonumber\\&\left.\frac{-\frac{\partial}{\partial w_{j}}\left(\exp\left(\sum_{i}\text{\ensuremath{y^{n}}}w_{i}x_{i}^{n}-\left\Vert {\bm w}\right\Vert \left\Vert {\bm \eta}^{n}\right\Vert \right)\right)}{\left(1+\exp\left(\sum_{i}\text{\ensuremath{y^{n}}}w_{i}x_{i}^{n}-\left\Vert {\bm w}\right\Vert \left\Vert {\bm \eta}^{n}\right\Vert \right)\right)^{2}}\right)\nonumber\\
&=\frac{1}{N}\sum_{n}\left(\frac{\frac{\left\Vert {\bm \eta}^{n}\right\Vert }{\left\Vert {\bm w}\right\Vert }\left(\delta_{ij}-\frac{w_{i}w_{j}}{\left\Vert {\bm w}\right\Vert ^{2}}\right)}{\left(1+\exp\left(\sum_{i}\text{\ensuremath{y^{n}}}w_{i}x_{i}^{n}-\left\Vert {\bm w}\right\Vert \left\Vert {\bm \eta}^{n}\right\Vert \right)\right)}\right.\nonumber\\&+\left(-\text{\ensuremath{y^{n}}}x_{i}^{n}+\frac{w_{i}}{\left\Vert {\bm w}\right\Vert }\left\Vert {\bm \eta}^{n}\right\Vert \right)\nonumber\\&\frac{-\left(\exp\left(\sum_{i}\text{\ensuremath{y^{n}}}w_{i}x_{i}^{n}-\left\Vert {\bm w}\right\Vert \left\Vert {\bm \eta}^{n}\right\Vert \right)\right)}{\left(1+\exp\left(\sum_{i}\text{\ensuremath{y^{n}}}w_{i}x_{i}^{n}-\left\Vert {\bm w}\right\Vert \left\Vert {\bm \eta}^{n}\right\Vert \right)\right)^{2}}\nonumber
\end{align}
\begin{align}
&\left.\frac{\partial}{\partial w_{j}}\left(\sum_{i}\text{\ensuremath{y^{n}}}w_{i}x_{i}^{n}-\left\Vert {\bm w}\right\Vert \left\Vert {\bm \eta}^{n}\right\Vert \right)\right)\nonumber\\&=\frac{1}{N}\sum_{n}\left(\frac{\frac{\left\Vert {\bm \eta}^{n}\right\Vert }{\left\Vert {\bm w}\right\Vert }\left(\delta_{ij}-\frac{w_{i}w_{j}}{\left\Vert {\bm w}\right\Vert ^{2}}\right)}{\left(1+\exp\left(\sum_{i}\text{\ensuremath{y^{n}}}w_{i}x_{i}^{n}-\left\Vert {\bm w}\right\Vert \left\Vert {\bm \eta}^{n}\right\Vert \right)\right)}\right.\nonumber\\&+\left(-\text{\ensuremath{y^{n}}}x_{i}^{n}+\frac{w_{i}}{\left\Vert {\bm w}\right\Vert }\left\Vert {\bm \eta}^{n}\right\Vert \right)\nonumber\\&\frac{-\left(\exp\left(\sum_{i}\text{\ensuremath{y^{n}}}w_{i}x_{i}^{n}-\left\Vert {\bm w}\right\Vert \left\Vert {\bm \eta}^{n}\right\Vert \right)\right)}{\left(1+\exp\left(\sum_{i}\text{\ensuremath{y^{n}}}w_{i}x_{i}^{n}-\left\Vert {\bm w}\right\Vert \left\Vert {\bm \eta}^{n}\right\Vert \right)\right)^{2}}\nonumber\\&\left.\left(\text{\ensuremath{y^{n}}}x_{j}^{n}-\frac{w_{j}}{\left\Vert {\bm w}\right\Vert }\left\Vert {\bm \eta}^{n}\right\Vert \right)\right)\nonumber\\
&=\frac{1}{N}\sum_{n}\left(\frac{\frac{\left\Vert {\bm \eta}^{n}\right\Vert }{\left\Vert {\bm w}\right\Vert }\left(\delta_{ij}-\frac{w_{i}w_{j}}{\left\Vert {\bm w}\right\Vert ^{2}}\right)}{\left(1+\exp\left(\sum_{i}\text{\ensuremath{y^{n}}}w_{i}x_{i}^{n}-\left\Vert {\bm w}\right\Vert \left\Vert {\bm \eta}^{n}\right\Vert \right)\right)}\right.\nonumber\\&+\frac{\exp\left(\sum_{i}\text{\ensuremath{y^{n}}}w_{i}x_{i}^{n}-\left\Vert {\bm w}\right\Vert \left\Vert {\bm \eta}^{n}\right\Vert \right)}{\left(1+\exp\left(\sum_{i}\text{\ensuremath{y^{n}}}w_{i}x_{i}^{n}-\left\Vert {\bm w}\right\Vert \left\Vert {\bm \eta}^{n}\right\Vert \right)\right)^{2}}\nonumber\\&\left.\left(\text{\ensuremath{y^{n}}}x_{i}^{n}-\frac{w_{i}}{\left\Vert {\bm w}\right\Vert }\left\Vert {\bm \eta}^{n}\right\Vert \right)\left(\text{\ensuremath{y^{n}}}x_{j}^{n}-\frac{w_{j}}{\left\Vert {\bm w}\right\Vert }\left\Vert {\bm \eta}^{n}\right\Vert \right)\right)\nonumber.
\end{align}
The optimal weight condition $\frac{w_{i}^{\ast}}{\left\Vert {\bm w}^{\ast}\right\Vert }=\frac{y^{n}x_{i}^{n}}{\left\Vert {\bm \eta}^{n}\right\Vert }$ eliminate the second terms. The first term is
\begin{align}
&\left.\frac{\partial^{2}L_{\eta}}{\partial\omega_{i}\partial\omega_{j}}\right|_{{\bm w}={\bm w}^{\ast}}\nonumber\\&=\frac{1}{N}\sum_{n}\frac{\frac{\left\Vert {\bm \eta}^{n}\right\Vert }{\left\Vert {\bm w}^{\ast}\right\Vert }\left(\delta_{ij}-\frac{w_{i}^{\ast}w_{j}^{\ast}}{\left\Vert {\bm w}^{\ast}\right\Vert ^{2}}\right)}{\left(1+\exp\left(\left(\sum_{i}\text{\ensuremath{y^{n}}}w_{i}^{\ast}x_{i}^{n}-\left\Vert {\bm w}^{\ast}\right\Vert \left\Vert {\bm \eta}^{n}\right\Vert \right)\right)\right)}\nonumber\\&=\frac{1}{N}\sum_{n}\frac{\frac{\left\Vert {\bm \eta}^{n}\right\Vert }{\left\Vert {\bm w}^{\ast}\right\Vert }\left(\delta_{ij}-\frac{w_{i}^{\ast}w_{j}^{\ast}}{\left\Vert {\bm w}^{\ast}\right\Vert ^{2}}\right)}{\left(1+\exp\left(\left(\sum_{i}w_{i}^{\ast}\frac{\left\Vert {\bm \eta}^{n}\right\Vert }{\left\Vert {\bm w}^{\ast}\right\Vert }w_{i}^{\ast}-\left\Vert {\bm w}^{\ast}\right\Vert \left\Vert {\bm \eta}^{n}\right\Vert \right)\right)\right)}\nonumber\\&=\frac{1}{N}\sum_{n}\frac{\frac{\left\Vert {\bm \eta}^{n}\right\Vert }{\left\Vert {\bm w}^{\ast}\right\Vert }\left(\delta_{ij}-\frac{w_{i}^{\ast}w_{j}^{\ast}}{\left\Vert {\bm w}^{\ast}\right\Vert ^{2}}\right)}{\left(1+\exp\left(\text{\ensuremath{y^{n}}}\left\Vert {\bm \eta}^{n}\right\Vert \left(\frac{\left\Vert {\bm w}^{\ast}\right\Vert ^{2}}{\left\Vert {\bm w}^{\ast}\right\Vert }-\left\Vert {\bm w}^{\ast}\right\Vert \right)\right)\right)}\nonumber\\&=\frac{1}{2N}\sum_{n}\frac{\left\Vert {\bm \eta}^{n}\right\Vert }{\left\Vert {\bm w}^{\ast}\right\Vert }\left(\delta_{ij}-\frac{w_{i}^{\ast}w_{j}^{\ast}}{\left\Vert {\bm w}^{\ast}\right\Vert ^{2}}\right)\nonumber.
\end{align}

\section{Experimental Details}
In this chapter we describe the experimental details. We use five image datasets: MNIST~\cite{lecun1998gradient}, MNIST2, CIFAR10~\cite{krizhevsky2009learning}, CIFAR2, and SVHN~\cite{netzer2011reading}. MNIST2 and CIFAR2 are the original datasets for binary classification. We made MNIST2 using only MNIST class 0 and class 1, and CIFAR2 using only frog and ship classes in CIFAR10. We chose the frog and ship classes because they have the highest classification accuracy. In CIFAR10 and CIFAR2, model architecture, data standardization, and the parameters of the PGD attack were set the same as in~\cite{madry2017towards}. In SVHN, model architecture, data standardization, and the parameters of the PGD attack were set the same as in~\cite{wu2020adversarial}.

\section{Additional Experiments}

\subsection{MNIST2 with the larger magnitude of noise}

\begin{figure}[tbh]
\centering
\includegraphics[width=0.6\linewidth]{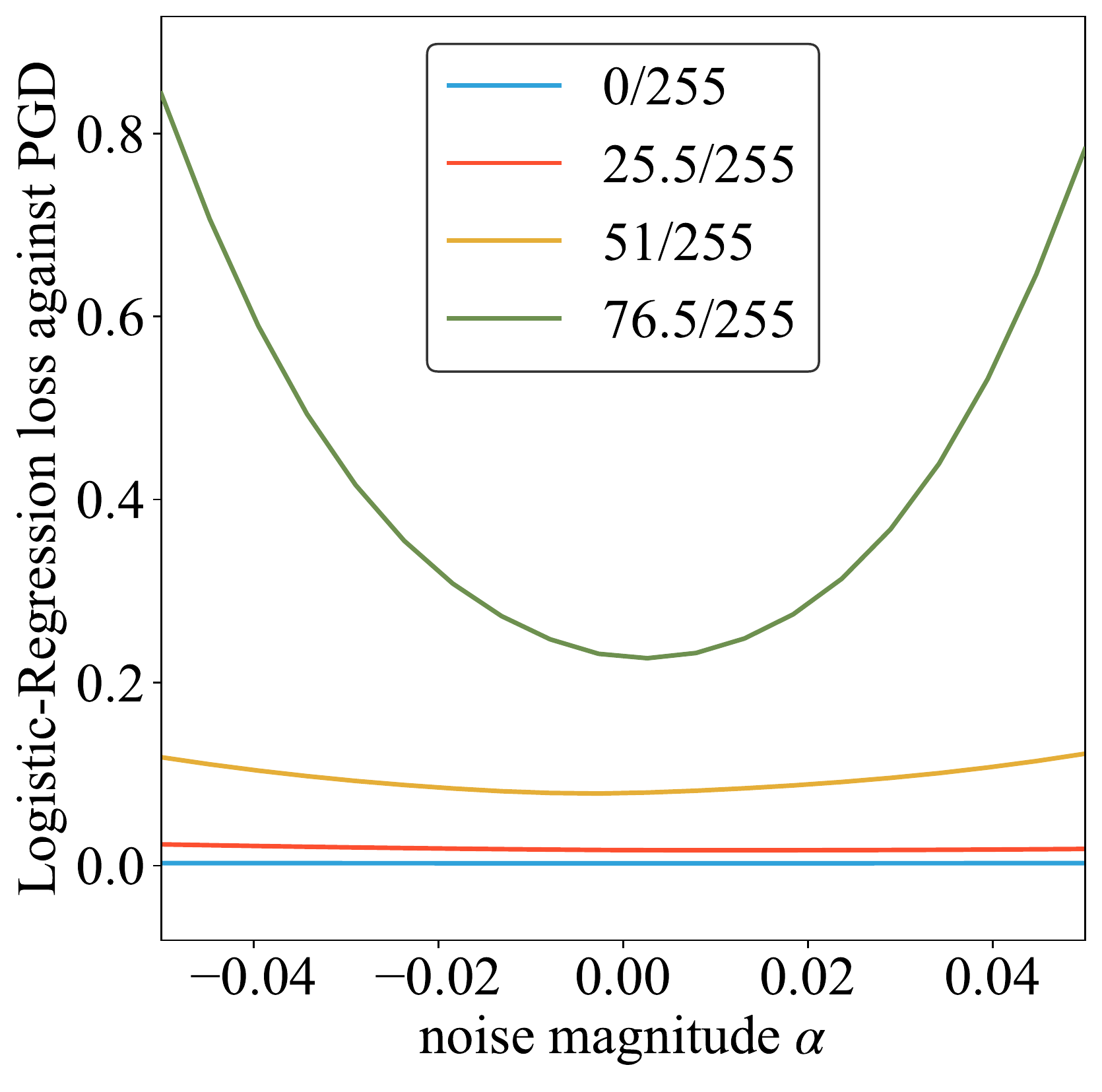}
\caption{Weight Loss Landscape against $L_{\infty}$PGD in MNIST2}
\label{fig:landscape-mnist2-large}
\end{figure}


Figure~\ref{fig:landscape-mnist2-large} shows the results of the experiment using the noise magnitude range $\varepsilon=\left\{ 0,\frac{25.5}{255},\frac{51}{255},\frac{76.5}{255}\right\}$, which is commonly used in MNIST for 10-class classification.
\bibliographystyle{named}
\bibliography{landscape}

\begin{thebibliography}{}

\bibitem[\protect\citeauthoryear{Chaudhari \bgroup \em et al.\egroup
  }{2017}]{chaudhari2017entropy}
Pratik Chaudhari, Anna Choromanska, Stefano Soatto, Yann LeCun, Carlo Baldassi,
  Christian Borgs, Jennifer Chayes, Levent Sagun, and Riccardo Zecchina.
\newblock Entropy-sgd: Biasing gradient descent into wide valleys.
\newblock In {\em ICLR}, 2017.

\bibitem[\protect\citeauthoryear{Cohen \bgroup \em et al.\egroup
  }{2019}]{cohen2019certified}
Jeremy Cohen, Elan Rosenfeld, and Zico Kolter.
\newblock Certified adversarial robustness via randomized smoothing.
\newblock In {\em ICML}, pages 1310--1320, 2019.

\bibitem[\protect\citeauthoryear{Croce and Hein}{2020}]{croce2020reliable}
Francesco Croce and Matthias Hein.
\newblock Reliable evaluation of adversarial robustness with an ensemble of
  diverse parameter-free attacks.
\newblock {\em ICML}, 2020.

\bibitem[\protect\citeauthoryear{Devlin \bgroup \em et al.\egroup
  }{2019}]{devlin2019bert}
Jacob Devlin, Ming-Wei Chang, Kenton Lee, and Kristina Toutanova.
\newblock Bert: Pre-training of deep bidirectional transformers for language
  understanding.
\newblock In {\em NAACL}, 2019.

\bibitem[\protect\citeauthoryear{Dinh \bgroup \em et al.\egroup
  }{2017}]{dinh2017sharp}
Laurent Dinh, Razvan Pascanu, Samy Bengio, and Yoshua Bengio.
\newblock Sharp minima can generalize for deep nets.
\newblock In {\em ICML}, 2017.

\bibitem[\protect\citeauthoryear{Foret \bgroup \em et al.\egroup
  }{2020}]{foret2020sharpness}
Pierre Foret, Ariel Kleiner, Hossein Mobahi, and Behnam Neyshabur.
\newblock Sharpness-aware minimization for efficiently improving
  generalization.
\newblock {\em arXiv preprint arXiv:2010.01412}, 2020.

\bibitem[\protect\citeauthoryear{Goodfellow \bgroup \em et al.\egroup
  }{2015}]{goodfellow2014explaining}
Ian Goodfellow, Jonathon Shlens, and Christian Szegedy.
\newblock Explaining and harnessing adversarial examples.
\newblock In {\em ICLR}, 2015.

\bibitem[\protect\citeauthoryear{He \bgroup \em et al.\egroup
  }{2016}]{he2016deep}
Kaiming He, Xiangyu Zhang, Shaoqing Ren, and Jian Sun.
\newblock Deep residual learning for image recognition.
\newblock In {\em CVPR}, pages 770--778, 2016.

\bibitem[\protect\citeauthoryear{Hochreiter and
  Schmidhuber}{1997}]{hochreiter1997flat}
Sepp Hochreiter and J{\"u}rgen Schmidhuber.
\newblock Flat minima.
\newblock {\em Neural Computation}, 9(1):1--42, 1997.

\bibitem[\protect\citeauthoryear{Hu \bgroup \em et al.\egroup
  }{2019}]{NEURIPS2019_cbb6a3b8}
Shengyuan Hu, Tao Yu, Chuan Guo, Wei-Lun Chao, and Kilian~Q Weinberger.
\newblock A new defense against adversarial images: Turning a weakness into a
  strength.
\newblock In {\em NeurIPS}, volume~32, pages 1635--1646, 2019.

\bibitem[\protect\citeauthoryear{Huang \bgroup \em et al.\egroup
  }{2019}]{huang2019model}
Bo~Huang, Yi~Wang, and Wei Wang.
\newblock Model-agnostic adversarial detection by random perturbations.
\newblock In {\em IJCAI}, pages 4689--4696, 2019.

\bibitem[\protect\citeauthoryear{Jiang \bgroup \em et al.\egroup
  }{2020}]{Jiang*2020Fantastic}
Yiding Jiang, Behnam Neyshabur, Hossein Mobahi, Dilip Krishnan, and Samy
  Bengio.
\newblock Fantastic generalization measures and where to find them.
\newblock In {\em ICLR}, 2020.

\bibitem[\protect\citeauthoryear{Keskar \bgroup \em et al.\egroup
  }{2017}]{keskar2016large}
Nitish~Shirish Keskar, Dheevatsa Mudigere, Jorge Nocedal, Mikhail Smelyanskiy,
  and Ping Tak~Peter Tang.
\newblock On large-batch training for deep learning: Generalization gap and
  sharp minima.
\newblock In {\em ICLR}, 2017.

\bibitem[\protect\citeauthoryear{Krizhevsky \bgroup \em et al.\egroup
  }{2009}]{krizhevsky2009learning}
Alex Krizhevsky, Geoffrey Hinton, et~al.
\newblock Learning multiple layers of features from tiny images.
\newblock {\em Technical report}, 2009.

\bibitem[\protect\citeauthoryear{Kurakin \bgroup \em et al.\egroup
  }{2017}]{kurakin2016adversarial}
Alexey Kurakin, Ian Goodfellow, and Samy Bengio.
\newblock Adversarial machine learning at scale.
\newblock In {\em ICLR}, 2017.

\bibitem[\protect\citeauthoryear{LeCun \bgroup \em et al.\egroup
  }{1998}]{lecun1998gradient}
Yann LeCun, L{\'e}on Bottou, Yoshua Bengio, and Patrick Haffner.
\newblock Gradient-based learning applied to document recognition.
\newblock {\em Proceedings of the IEEE}, 1998.

\bibitem[\protect\citeauthoryear{Li \bgroup \em et al.\egroup
  }{2018}]{li2018visualizing}
Hao Li, Zheng Xu, Gavin Taylor, Christoph Studer, and Tom Goldstein.
\newblock Visualizing the loss landscape of neural nets.
\newblock In {\em NeurIPS}, pages 6389--6399, 2018.

\bibitem[\protect\citeauthoryear{Liu \bgroup \em et al.\egroup
  }{2019}]{liu2019detection}
Jiayang Liu, Weiming Zhang, Yiwei Zhang, Dongdong Hou, Yujia Liu, Hongyue Zha,
  and Nenghai Yu.
\newblock Detection based defense against adversarial examples from the
  steganalysis point of view.
\newblock In {\em CVPR}, pages 4825--4834, 2019.

\bibitem[\protect\citeauthoryear{Liu \bgroup \em et al.\egroup
  }{2020}]{liu2020loss}
Chen Liu, Mathieu Salzmann, Tao Lin, Ryota Tomioka, and Sabine S{\"u}sstrunk.
\newblock On the loss landscape of adversarial training: Identifying challenges
  and how to overcome them.
\newblock {\em NeurIPS}, 33, 2020.

\bibitem[\protect\citeauthoryear{Madry \bgroup \em et al.\egroup
  }{2018}]{madry2017towards}
Aleksander Madry, Aleksandar Makelov, Ludwig Schmidt, Dimitris Tsipras, and
  Adrian Vladu.
\newblock Towards deep learning models resistant to adversarial attacks.
\newblock In {\em ICLR}, 2018.

\bibitem[\protect\citeauthoryear{Metzen \bgroup \em et al.\egroup
  }{2017}]{metzen2017detecting}
Jan~Hendrik Metzen, Tim Genewein, Volker Fischer, and Bastian Bischoff.
\newblock On detecting adversarial perturbations.
\newblock {\em ICLR}, 2017.

\bibitem[\protect\citeauthoryear{Mobahi}{2016}]{Mobahi2016}
Hossein Mobahi.
\newblock Training recurrent neural networks by diffusion.
\newblock {\em CoRR}, abs/1601.04114, 2016.

\bibitem[\protect\citeauthoryear{Netzer \bgroup \em et al.\egroup
  }{2011}]{netzer2011reading}
Yuval Netzer, Tao Wang, Adam Coates, Alessandro Bissacco, Bo~Wu, and Andrew~Y
  Ng.
\newblock Reading digits in natural images with unsupervised feature learning.
\newblock {\em In NeurIPS Workshop on Deep Learning and Unsupervised Feature
  Learning}, 2011.

\bibitem[\protect\citeauthoryear{Papernot \bgroup \em et al.\egroup
  }{2016}]{papernot2016distillation}
Nicolas Papernot, Patrick McDaniel, Xi~Wu, Somesh Jha, and Ananthram Swami.
\newblock Distillation as a defense to adversarial perturbations against deep
  neural networks.
\newblock In {\em S\&P (Oakland)}, pages 582--597. IEEE, 2016.

\bibitem[\protect\citeauthoryear{Prabhu \bgroup \em et al.\egroup
  }{2019}]{prabhu2019understanding}
Vinay~Uday Prabhu, Dian~Ang Yap, Joyce Xu, and John Whaley.
\newblock Understanding adversarial robustness through loss landscape
  geometries.
\newblock {\em arXiv preprint arXiv:1907.09061}, 2019.

\bibitem[\protect\citeauthoryear{Rice \bgroup \em et al.\egroup
  }{2020}]{rice2020overfitting}
Leslie Rice, Eric Wong, and Zico Kolter.
\newblock Overfitting in adversarially robust deep learning.
\newblock In {\em ICML}, pages 8093--8104. PMLR, 2020.

\bibitem[\protect\citeauthoryear{Salman \bgroup \em et al.\egroup
  }{2019}]{salman2019provably}
Hadi Salman, Jerry Li, Ilya Razenshteyn, Pengchuan Zhang, Huan Zhang, Sebastien
  Bubeck, and Greg Yang.
\newblock Provably robust deep learning via adversarially trained smoothed
  classifiers.
\newblock In {\em NeurIPS}, pages 11292--11303, 2019.

\bibitem[\protect\citeauthoryear{Szegedy \bgroup \em et al.\egroup
  }{2014}]{szegedy2013intriguing}
Christian Szegedy, Wojciech Zaremba, Ilya Sutskever, Joan Bruna, Dumitru Erhan,
  Ian Goodfellow, and Rob Fergus.
\newblock Intriguing properties of neural networks.
\newblock In {\em ICLR}, 2014.

\bibitem[\protect\citeauthoryear{Wang \bgroup \em et al.\egroup
  }{2017}]{wang2017residual}
Yisen Wang, Xuejiao Deng, Songbai Pu, and Zhiheng Huang.
\newblock Residual convolutional ctc networks for automatic speech recognition.
\newblock {\em arXiv preprint arXiv:1702.07793}, 2017.

\bibitem[\protect\citeauthoryear{Wang \bgroup \em et al.\egroup
  }{2020}]{Wang2020Improving}
Yisen Wang, Difan Zou, Jinfeng Yi, James Bailey, Xingjun Ma, and Quanquan Gu.
\newblock Improving adversarial robustness requires revisiting misclassified
  examples.
\newblock In {\em ICLR}, 2020.

\bibitem[\protect\citeauthoryear{Wu \bgroup \em et al.\egroup
  }{2020}]{wu2020adversarial}
Dongxian Wu, Shu-Tao Xia, and Yisen Wang.
\newblock Adversarial weight perturbation helps robust generalization.
\newblock {\em NeurIPS}, 33, 2020.

\bibitem[\protect\citeauthoryear{Yang \bgroup \em et al.\egroup
  }{2020}]{pmlr-v119-yang20c}
Greg Yang, Tony Duan, J.~Edward Hu, Hadi Salman, Ilya Razenshteyn, and Jerry
  Li.
\newblock Randomized smoothing of all shapes and sizes.
\newblock In {\em ICML}, volume 119, pages 10693--10705, 2020.

\bibitem[\protect\citeauthoryear{Zhang \bgroup \em et al.\egroup
  }{2019}]{zhang2019theoretically}
Hongyang Zhang, Yaodong Yu, Jiantao Jiao, Eric~P. Xing, Laurent~El Ghaoui, and
  Michael~I. Jordan.
\newblock Theoretically principled trade-off between robustness and accuracy.
\newblock In {\em ICML}, 2019.

\end{thebibliography}

\end{document}